\documentclass[letterpaper, 10 pt, conference]{ieeeconf}  %
\IEEEoverridecommandlockouts                              %

\newcommand{\calI}{\mathcal{I}}

\newcommand{\calM}{\mathcal{M}}

\newcommand{\calT}{\mathcal{T}}

\newcommand{\calX}{\mathcal{X}}

\newcommand\x{{\boldsymbol{\theta}}}

\newcommand{\inner}[2]{\langle {#1}, {#2} \rangle}

\newcommand\thet{{\boldsymbol{\theta}}}

\newcommand\xs{{\boldsymbol{\x}^*}}

\newcommand\ok{o_t}

\newcommand\okn{o_{i,k}}
\newcommand{\E}[1]{\mathbb{E}\left[#1\right]}

\newcommand{\tmix}{\tau_{mix}}
\newcommand{\tmax}{\tau_{max}}

\newcommand{\bfg}{\mathbf{g}}
\newcommand{\barg}{\bar{\mathbf{g}}}

\newcommand{\bO}[1]{O\left(#1\right)}
\newcommand{\eqal}[2]{\begin{equation}\begin{aligned}\label{#1}
			#2
\end{aligned}\end{equation}}

\newcommand\mudiktau{\eta_{i, t_{i,k}}}
\newcommand\mudjktau{\eta_{j, t_{j,k}}}

\newcommand\vk{\mathbf{v}_k}

\newcommand{\odit}{o_{i, t_{i, k}}}
\newcommand{\odjt}{o_{j, t_{j, k}}}
\newcommand{\dgit}{\mathbf{g}(\x_{t_{i, k}}, o_{i, t_{i, k}})}

\newcommand{\pars}[1]{\left( {#1} \right)}
\usepackage{amsthm} 
\usepackage{amssymb}  
\usepackage{breqn}

\newtheorem{theorem}{Theorem}

\newtheorem{lemma}{Lemma}
\newtheorem{remark}{Remark}
\newtheorem{assumption}{Assumption}
\newtheorem{definition}{Definition}

\usepackage{amsmath,amsfonts,amssymb,color}
\usepackage{mathtools}

\usepackage{epsfig}
\usepackage{algorithm}
\usepackage{algpseudocode}
\usepackage{dsfont}
\usepackage{multirow}

\usepackage{array}
\usepackage{multirow}
\usepackage{mathtools}

\usepackage{epstopdf}
\usepackage{tikz}
\usepackage{relsize}
\usetikzlibrary{shapes,arrows}
\usepackage{cite}
\usepackage{epstopdf}
\usepackage{tikz}
\usetikzlibrary{shapes}
\usepackage[labelfont=normalfont]{caption}

\definecolor{winered}{rgb}{0.5,0,0}
\usepackage{scalerel}
\usepackage{xcolor}
\usepackage[colorlinks]{hyperref}
\AtBeginDocument{%
  \hypersetup{
    citecolor=blue,
    linkcolor=winered}}

\title{\LARGE \bf
\texttt{DASA}: Delay-Adaptive Multi-Agent Stochastic Approximation 
}

\author{Nicol\`o Dal Fabbro$^*$, Arman Adibi$^*$, H. Vincent Poor, Sanjeev R. Kulkarni, Aritra Mitra and George J. Pappas
	\thanks{
 N. Dal Fabbro and G. J. Pappas are with the Electrical and Systems Engineering Department, University of Pennsylvania. A. Adibi, H. V. Poor, and S. R. Kulkarni, are with the Department of Electrical and Computer Engineering, Princeton University. A. Mitra is with the Electrical and Computer Engineering Department, North Carolina State University. }
}

\begin{document}

\maketitle
\def\thefootnote{*}\footnotetext{These authors contributed equally to this work.}\def\thefootnote{\arabic{footnote}}

\thispagestyle{empty}
\pagestyle{empty}

\begin{abstract}
We consider a setting in which $N$ agents aim to speedup a common Stochastic Approximation (SA) problem by acting in parallel and communicating with a central server. We assume that the up-link transmissions to the server are subject to asynchronous and potentially unbounded time-varying delays. To mitigate the effect of delays and stragglers while reaping the benefits of distributed computation, we propose \texttt{DASA}, a Delay-Adaptive algorithm for multi-agent Stochastic Approximation. We provide a finite-time analysis of \texttt{DASA} assuming that the agents' stochastic observation processes are independent Markov chains. Significantly advancing existing results, \texttt{DASA} is the first algorithm whose convergence rate depends only on the mixing time $\tmix$ and on the average delay $\tau_{avg}$ while jointly achieving an $N$-fold convergence speedup under Markovian sampling. Our work is relevant for various SA applications, including multi-agent and distributed temporal difference (TD) learning, Q-learning and stochastic optimization with correlated data.

\end{abstract}
\section{Introduction}
Motivated by emerging applications in data-hungry algorithms like distributed large-scale reinforcement learning (RL), we study an instance of Stochastic Approximation (SA) in which multiple agents act in parallel to cooperatively solve a common SA problem. The main goal of splitting the computational effort across $N$ agents is to achieve an $N$-fold convergence speedup in the sample complexity relative to when a single agent acts alone. Only recently, theoretical studies have shown the beneficial effect of distributed computing in multi-agent and federated SA when agents' observations are temporally correlated (Markov) processes~\cite{khodadadian, dal2023federated, zhang2024finite}. However, in real-world large-scale communication networks and client-server architectures, inter-node communications are commonly subject to (potentially large) delays. 
While several algorithms have been provided to study and mitigate the effect of delays in distributed asynchronous optimization and federated learning~\cite{bertsekas1989convergence,koloskova2022sharper, stich2020error,adibi2023min} under i.i.d. sampling, very little is known about distributed SA under Markovian sampling. Existing works, indeed, either provide algorithms that are heavily impacted by the maximum delay~\cite{AMARL_th, feyzmahdavian2014delayed} (which can be potentially very large), or are not able to provide convergence speedups under Markovian sampling~\cite{adibi2024stochastic}. In the recent work in~\cite{AMARL_th}, the authors study collaborative gains of parallel learners in RL under Markovian sampling, but they obtain convergence rates affected by the maximum network delay and provide convergence speedups only when the data are collected in an i.i.d. fashion. However, note that the main challenges in proving multi-agent convergence speedups of SA algorithms emerge when agents' samples form temporally correlated trajectories. Indeed, even for single-agent settings, non-asymptotic convergence rates for SA under Markovian sampling have only recently been established~\cite{bhandari2018, srikant2019finite, chen2022finite}. In our recent work~\cite{adibi2024stochastic}, we provide finite-time convergence rates for an SA setting under Markovian sampling in which the SA operator is computed with delayed updates, and we propose a delay-adaptive algorithm that replaces the dependence on the maximum delay $\tmax$ with a dependence on the average delay $\tau_{avg}$. However, the work in~\cite{adibi2024stochastic} only considers a single-agent setting. In this paper, we bridge this gap proposing and analyzing a new multi-agent algorithm for distributed asynchronous SA, \texttt{DASA}. The main contributions of this paper are as follows.

\textbf{Contributions:}
We propose and provide a finite-time analysis of \texttt{DASA}, that has the following unprecedented features: (i) \texttt{DASA} has a convergence rate that depends on the network delays only via the average delay $\tau_{avg}$ and (ii) at the same time, \texttt{DASA} achieves an $N$-fold linear convergence speedup with the number of agents under Markovian sampling. Providing \textit{jointly} (i) and (ii) presents major technical challenges that we carefully address in our analysis in Section~\ref{sec:analysis}-~\ref{sec:proof}.
Note that we do not need conditions on the maximum delay: it can be as large as the global number of iterations $T$ and our results would still hold. 
To validate our theoretical findings, we provide simple simulations on a distributed TD learning problem in Section~\ref{sec:sim}.
\begin{figure}
\includegraphics[width=0.49\textwidth]{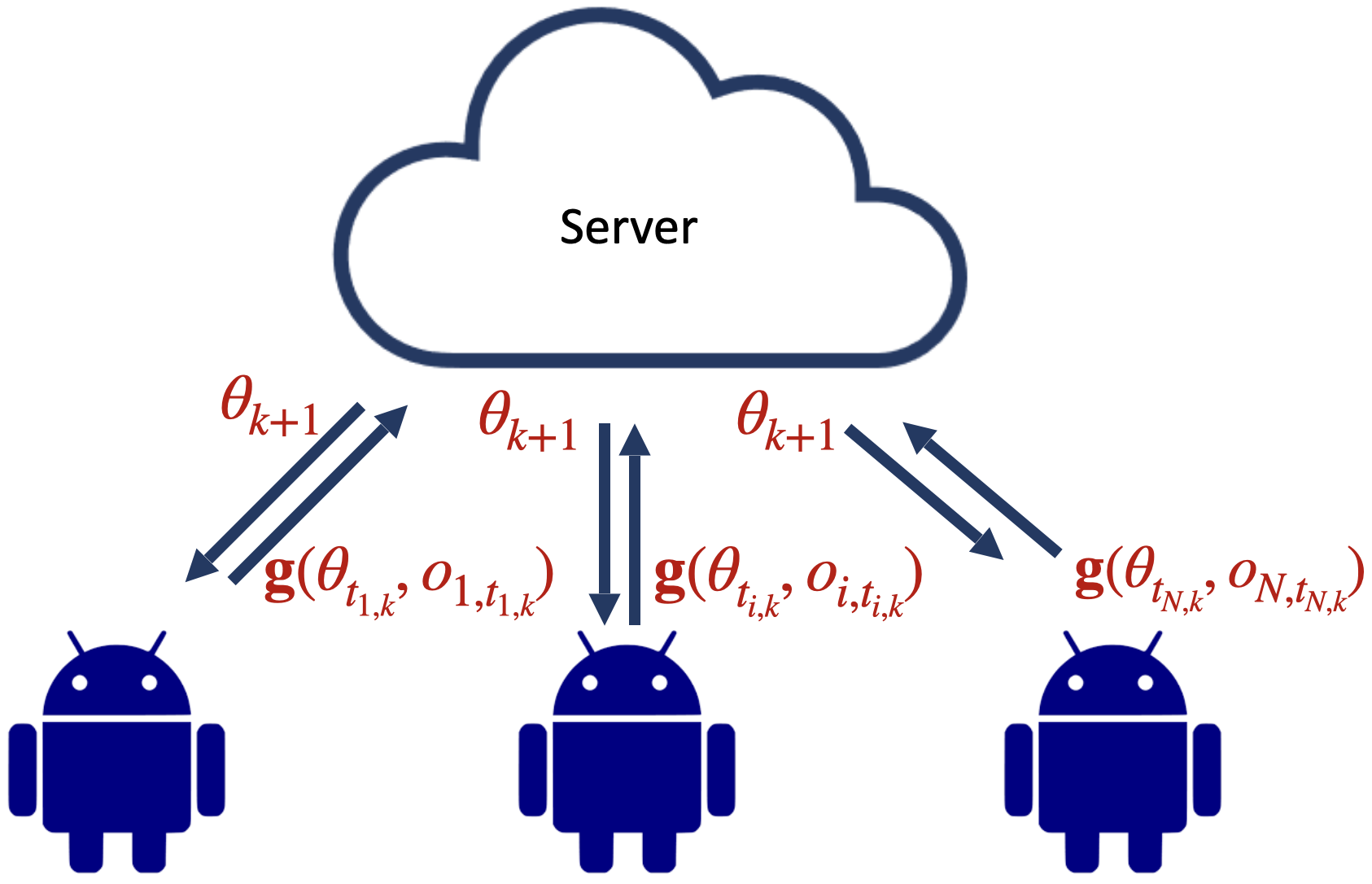}
    \caption{\texttt{DASA} system model. Agents $1, \dots, N$ cooperatively work to solve the SA problem. At iteration $k$, the server updates $\thet_k$ using delayed agents' operators $\{\mathbf{g}(\thet_{t_{i,k}}, o_{i, t_{i,k}})\}_{i = 1}^N$ and broadcasts $\thet_{k+1}$ to the agents.}
    \label{fig:setting}
\end{figure}

\section{Problem Setting and \texttt{DASA}}\label{sec:setting}
The goal of SA is to solve a root-finding problem formulated as follows:
\begin{equation}\label{eq:SA_problem}
    \textrm{Find } \xs \in \mathbb{R}^m \textrm{ such that } \barg(\xs) = 0, \hspace{0.2cm} 
\end{equation}
where, for a given approximation parameter ${\thet} \in \mathbb{R}^m$, the deterministic function $\barg(\thet)$ represents the expectation of a noisy operator $\mathbf{g}(\thet, o_k)$ taken over distribution $\pi$, and \(\{o_k\}\) denotes a stochastic observation process. It is commonly assumed that $\{o_k\}$ converges in distribution to $\pi$~\cite{borkar2009stochastic, meyn2023stability}.

\textbf{Distributed Asynchronous SA.} We consider a setting in which $N$ agents communicate with a central server to collaboratively estimate $\thet^*$. To achieve this goal, the agents compute in parallel their versions of the noisy operator $\bfg(\thet_k, o_{i,k})$, where $\thet_k$ is the approximation parameter at iteration $k$ and $\{o_{i,k}\}_{k\geq0}$ is the local observation process of agent  $i$. At each iteration, the agents transmit to a server their local operators via up-link communications subject to asynchronous time-varying delays. We denote by $\tau_{i,k}$ the delay of agent $i$ at iteration $k$. At each iteration $k$, the server can use the (delayed) update directions $\{\mathbf{g}(\thet_{t_{i,k}}, o_{i, t_{i,k}})\}_{i = 1}^N$ to update the SA parameter $\thet_k$, where $t_{i,k} = k-\tau_{i,k}$ is the iteration at which the SA update direction of agent $i$ was computed. On the other hand, \textit{we assume that the down-link communication from the server to the agents is not subject to delays}. This assumption is practically motivated by the fact that in most client-server architectures the main communication bottleneck comes from up-link transmissions, rather than from down-link broadcasting. \textit{We anticipate that this assumption can be removed and we plan to do so in future work.} The aggregation protocol is illustrated in Figure~\ref{fig:setting}. SA with delayed updates can estimate $\thet^*$, but with significant performance degradation in the rate{, as theoretically shown, e.g., in \cite{feyzmahdavian2014delayed, adibi2023min, adibi2024stochastic}} especially in the presence of large delays. To mitigate the impact of stragglers on the performance, we propose \texttt{DASA}, in which, using available delayed operators, the server updates the parameter $\thet_k$ according to a delay-adaptive update rule, which we describe next.

\textbf{\texttt{DASA}:} We now present \texttt{DASA}.
The server updates the parameter $\thet_k$ according to the following rule: 
\begin{equation}\label{eq:updateRule}
	\x_{k+1} = \x_k + \alpha\mathbf{v}_k,
\end{equation}
where $\alpha$ is a constant step-size/learning rate, and 

\begin{equation}
\begin{aligned}\label{eq:algo}
\mathbf{v}_k \triangleq
\begin{cases}
\frac{1}{|\calI_k|}\sum\limits_{i\in \calI_k}\mathbf{g}(\x_{t_{i,k}}, o_{i, t_{i,k}}) & \textrm{if}\hspace{0.1cm}\|\x_{t_{i_{med, k},k}}-\x_{k}\|\leq {\epsilon} \\
0  &\textrm{otherwise}
\end{cases},  
\end{aligned}
\end{equation}
where $\|\cdot\|$ denotes the Euclidean norm, while $i_{med, k}$, $\calI_k$ and $\epsilon$ are defined as follows:
\begin{equation}\label{eq:parameters}
\begin{aligned}
i_{med, k}&\triangleq\arg \text{median}(\{\|\x_{t_{j,k}}-\x_{k}\|\}_{j=1}^{N}),\\[0.3cm]
\calI_k&\triangleq\{i:\|\x_{t_{i,k}}-\x_{k}\|\leq\|\x_{t_{i_{med, k},k}}-\x_{k}\|\},\\[0.3cm]
\epsilon &\triangleq \max\{\alpha^{1.5}, \frac{\alpha}{\sqrt{M}}\},
\end{aligned}
\end{equation}
with $M = |\calI_k| = \lceil\frac{N}{2}\rceil, \forall k\geq 0$.
\begin{remark}
Note that the size of $\calI_k$ is $|\calI_k|=\lceil\frac{N}{2}\rceil$. Here, we assume without loss of generality that $N$ is even. Therefore, $|\mathcal{I}_k|=\frac{N}{2}$. The proof is equivalent for odd $N$.
\end{remark}
\begin{remark}
If $i\in \calI_k$, then we have 
\begin{align}
    \|\x_{t_{i,k}}-\x_{k}\|\leq \epsilon.
\end{align}
\end{remark}
The rationale behind \texttt{DASA} is to control the error introduced by delays, using, at each iteration, the information from the set of $M = N/2$ agents in $\calI_k$ whose available operators were computed with iterates that are \textit{the least stale at iteration $k$}, as dictated by the median delay error index $i_{med,k}$. At the same time, averaging ($N/2$) noisy operators, we wish to benefit from a variance reduction effect, achieving an $N$-fold convergence speedup thanks to the independence of the different stochastic observation processes. \texttt{DASA} allows the server to control the error of the aggregated operator $\vk$ that is used to update $\thet_k$, via the threshold parameter $\epsilon$. When carefully chosing $\epsilon$, we obtain the convergence guarantees that we want: \textit{a convergence rate that does not depend on the maximum delay and, jointly, a linear convergence speedup with the number of agents N}. Providing these convergence guarantees presents key technical challenges that we address in the next section.
\section{Convergence Analysis}\label{sec:analysis}
In this section, we provide the convergence analysis of \texttt{DASA}, under standard assumptions that we illustrate next. These assumptions are common in most finite-time analyses of SA algorithms~\cite{bhandari2018, srikant2019finite, chen2022finite}.
\begin{assumption}\label{ass:strongconvex}
Problem~\eqref{eq:SA_problem} admits a solution $\xs$, and $\exists \,\mu > 0$ such that for all $\thet\in {\mathbb{R}}^{m}$, we have
 \begin{equation}\label{eq:strongConvex}
{\langle\thet -\thet^*, \bar{\bfg}(\thet)-\bar{\bfg}(\thet^*)\rangle\leq -\mu \|\thet-\thet^*\|^2.}
 \end{equation}
\end{assumption}

\begin{assumption}\label{ass:Lipschitz}
	There exists $L >0$ such that for any $\thet_1, \thet_2\in\mathbb{R}^m$ and $o\in \{\ok\}$, we have
\begin{equation}
\label{eq:Lipschitz}
\|\bfg(\thet_1,o)-\bfg(\thet_2,o)\|\leq L\|\thet_1-\thet_2\|.
\end{equation}
 
 Furthermore, there exists $\sigma \geq 1$ such that for any $\thet\in\mathbb{R}^m$, we have 
 \eqal{eq:bGSq}{\|\bfg(\thet,o)\|\leq L(\|\thet\| + \sigma).}
\end{assumption}
\begin{assumption}
For each agent $i = 1, \dots, N$, the stochastic observation process $\{o_{i,k}\}$ is an aperiodic and irreducible Markov chain. For each $i\neq j$ and for each iteration $k$, $o_{i,k}$ and $o_{j,k}$ are statistically independent.
\label{assump:mixing}
\end{assumption}

Assumption \ref{ass:strongconvex} ensures the strong monotonicity of the expected operator $\barg(\cdot)$, which is a standard requirement for proving linear convergence rates.
Assumption \ref{ass:Lipschitz} imposes Lipschitz continuity on the noisy operator $\mathbf{g}(\cdot, \cdot)$, which is also common in SA analyses.
Assumption \ref{assump:mixing} implies that the Markov chain $\{o_k\}$ mixes at a geometric rate.

\begin{definition} \label{def:mix} 
{Let $\tmix$ be such that, for every $\boldsymbol{\theta} \in \mathbb{R}^m$, for all $i = 1, ..., N$ and time-step $\Bar{t}$, we have} 
\hspace{-10mm}
\begin{equation}
\begin{aligned}
\Vert \mathbb{E}\left[\mathbf{g}(\boldsymbol{\theta}, o_{i,t})|o_{i,\bar{t}}\right]-\barg(\boldsymbol{\theta})\Vert \leq \alpha^2 \left(\Vert \boldsymbol{\theta} \Vert +\sigma \right), &\\{\forall t: t-\bar{t} \geq \tmix}, \forall o_{i, \bar{t}}. &
\end{aligned}
\label{eqn:mix}
\end{equation}
\hspace{-4mm}
\end{definition}
Definition \ref{def:mix} introduces the mixing time $\tau_{mix}$, which quantifies the number of steps after which the conditional expectation of $\mathbf{g}(\thet, o_{i,t})$ is close to its stationary expectation $\mathbf{\bar{g}}(\thet)$, after conditioning over any past observation $o_{i,\bar{t}}$. Note that, thanks to Assumption~\ref{assump:mixing}, for any $i=1,...N$, the Markov chain $\{o_{i,k}\}$ mixes at a geometric rate, and hence there exists some $K\geq1$ such that $\tmix\leq K\log(\frac{1}{\alpha})$. 
Under the assumptions above, we are now ready to state the main result of this paper, which is the non-asymptotic convergence analysis of \texttt{DASA}. We provide the convergence result in the following theorem, whose proof requires a novel and careful analysis, which we defer to the next section. In all of our results, we use the notation  $y = O(x)$ to describe the relation between positive terms $x$ and $y$, meaning that the value $y$ is at most equal to some positive constant multiplying the quantity $x$. We define $\delta_k \triangleq \|\thet_k - \thet^*\|$, and assume without loss of generality that $\sigma \geq \max \{1, \|\thet^*\|, \|\thet_0\|\}$, $L\geq 1$ and $\mu \leq 1$. We also define an indicator function $\calX_k$, that is equal to 1 when the median agent's parameter is not too far away from the server's parameter $\thet_k$, and the average delay $\tau_{avg}$:
\begin{equation}
\begin{aligned}
\calX_k&\triangleq \begin{cases}
    1 & \textrm{if}\hspace{0.1cm}\|\x_{t_{i_{med, k},k}}-\x_{k}\|\leq {\epsilon}\\
    0 &\textrm{otherwise}
\end{cases},
\\
\tau_{avg}&\triangleq\frac{1}{NT}\sum_{t=1}^{T}\sum_{i=1}^{N}\tau_{i,t}.
\end{aligned}
\end{equation}
\begin{theorem}\label{th:main}
Consider the update rule of \texttt{DASA} in~\eqref{eq:updateRule} and let $T\geq 0$ be such that $\calX_T = 1$. There exists a universal constant $C_1\geq 1$ such that for $\alpha \leq \frac{\mu}{C_1L^2\tmix}$ the following holds:
\begin{equation}
\begin{aligned}
    \E{\delta_{T}^2}\leq\exp\left(\frac{-\alpha\mu T}{8\left(\tau_{avg} +1\right)}\right)\bO{\sigma^2} &\\ +\bO{\frac{\sigma^2L^2\alpha\tmix}{\mu}}\left(\frac{1}{N} + \alpha\right)&
\end{aligned}
\end{equation}
\end{theorem}
\textbf{Main Takeaways:} We now discuss the main takeaways from Theorem~\ref{th:main}. We first note that, with a constant step size $\alpha$ satisfying the theorem conditions, \texttt{DASA} guarantees mean-square exponential convergence of the approximation parameter $\thet_T$ to a ball around $\xs$. We now compare this result against existing related work and comment on the effect of the delays on the convergence rate. We also illustrate explicitly the linear convergence speedup achieved by \texttt{DASA}.

\textit{Effect of Delays and Dependence on $\tau_{avg}$.} Note that the exponential convergence rate of \texttt{DASA} depends on the average delay $\tau_{avg}$, while neither the rate nor the convergence ball are affected by the maximum delay $\tau_{max}$. The convergence ball, in fact, does not depend on the delay sequence at all. In this regard, note that convergence rates of non-adaptive SA algorithms under delayed updates are impacted by the maximum delay $\tau_{max}$ both in the convergence rate and in the noise ball of convergence, \textit{even in a single-agent setting}\cite{adibi2024stochastic}. At the same time, the step size choice that guarantees the convergence of \texttt{DASA} does not require any knowledge of the delay sequence, while it requires knowledge of the mixing time of the Markov sampling process, which is commonly required for finite-time convergence results in SA~\cite{srikant2019finite, bhandari2018, chen2022finite}. With respect to the single-agent SA setting with delayed updates in~\cite{adibi2024stochastic}, \texttt{DASA} manages to remove the dependence on the maximum delay $\tau_{max}$ from both the convergence rate and from the step size choice even in the multi-agent setting \textit{while jointly achieving a linear convergence speedup with the number of agents under Markovian sampling}. We now comment on the convergence speedup. 

\textit{Linear Convergence Speedup.} 
We now compare the convergence bound of \texttt{DASA} with the convergence bound obtained by the single-agent delay-adaptive algorithm provided in~\cite{adibi2024stochastic}. Note that the noise ball is deflated by $\frac{1}{N} + \alpha$, which, for a proper choice of step size $\alpha$ and for a number of iterations $T$ large enough, provides a linear convergence speedup with the number of agents $N$. In particular, we can easily show that if we set $\alpha$ and $T$ as follows:
\begin{equation*}
\alpha = \bO{\frac{\log(NT)\tau_{avg}}{T\mu}}, \hspace{0.2cm}T\geq \frac{CN\log(NT)\tau_{avg}\tmix L^2}{\mu^2},
\end{equation*}
with $C\geq 2$ a constant, then the mean-square error after $T$ iterations is
\begin{equation}
\E{\delta_T^2}\leq \bO{\frac{\log(NT)\tau_{avg}\tmix\sigma^2L^2}{NT\mu}},
\end{equation}
which implies that \texttt{DASA} is \textit{sample efficient} and that, despite the joint presence of asynchronous delays and Markovian sampling, we can achieve an $N$-fold linear convergence speedup when $N$ agents act in parallel, without dependence on the maximum delay. \textit{\texttt{DASA} is the first algorithm for multi-agent/distributed SA with these convergence guarantees.} 
\section{Proof of Theorem~\ref{th:main}}\label{sec:proof}
In this section, we prove Theorem~\ref{th:main}, providing the complete convergence analysis of \texttt{DASA}. We start by introducing some notation. Let us define the following:
\begin{equation}
\begin{aligned}
\mathcal{T}_k &\triangleq \{\tau_{i,k}: i\in\mathcal{I}_k\},\\[0.2cm]
\eta_{i,k}(\thet) &\triangleq \|\E{\mathbf{g}_{i,k}(\x, \okn)|o_{i, k-\tmix}} - \barg(\x)\|,\ k\geq \tmix, \\[0.2cm]
\delta_{k, h}^2& \triangleq \|\thet_k - \thet_{k-h}\|^2, k\geq h\geq 0.
\end{aligned}
\end{equation}
\textbf{Outline of the Analysis:} As a first step, we here provide a high-level outline of the convergence analysis of \texttt{DASA}. We start by writing the following recursion, which holds for the iterations $k\geq\tmix$ in which $\calX_k = 1$,
\begin{equation}\label{eq:outRecursion}
\begin{aligned}
\E{\delta_{k+1}^2} &= \E{\delta_k^2} - 2\alpha\E{\inner{\vk}{\thet_k - \thet^*}} + \alpha^2\E{\|\vk\|^2}\\
&= \E{\delta_k^2} -2\alpha\E{\inner{\barg(\thet_k)}{\thet_k-\thet^*}}\\
&-2\alpha\E{\psi_k} + \alpha^2\E{\|\vk\|^2},\\
&\overset{(*)}{\leq} (1-2\alpha\mu)\E{\delta_k^2} \\& -2\alpha\E{\psi_k} + \alpha^2\E{\|\vk\|^2},
\hspace{0.2cm}
\textrm{with}\\
\psi_k &\triangleq \inner{\vk - \barg(\thet_k)}{\thet_k - \thet^*},
\end{aligned}
\end{equation}
where for inequality $(*)$ we used Assumption~\ref{ass:strongconvex}. The main objective of the analysis is to obtain a one-step bound of the following form (for iterations $k$ in which an update occurs, i.e., in which $\calX_k = 1$):
\begin{equation}\label{eq:outOneStep}
\begin{aligned}
\E{\delta_{k+1}^2}&\leq \exp\left(-\alpha\mu\right)\E{\delta_k^2} \\&+ \bO{\sigma^2L^2\alpha^2\tmix}\left(\frac{1}{N}+\alpha\right),
\end{aligned}
\end{equation}
where we note that there is not any dependence on the maximum delay $\tmax$. In order to obtain this one-step bound, we need to provide bounds on the two terms $\E{\|\vk\|^2}$ and $\E{\psi_k}$ of the following form:
\begin{equation}\label{eq:outDesired}
\begin{aligned}
\E{\|\vk\|^2}&\leq \bO{L^2}\E{\delta_k^2} + \bO{L^2\frac{\sigma^2}{N}}&&\\
&+ \bO{\sigma^2\alpha^4} + \bO{L^2\epsilon^2},&&
\\
\E{\psi_k}&\leq \bO{\alpha\tmix L^2}\E{\delta_k^2} &&\\
&+\bO{\alpha\tmix} \left(L^2\frac{\sigma^2}{N}+\epsilon^2L^2 +\sigma^2\alpha^4 \right). 
\end{aligned}
\end{equation}
Providing these two bounds is the main technical burden of the proof of Theorem~\ref{th:main}. The bound on $\E{\|\vk\|^2}$ is the first key result that is used then to bound $\E{\psi_k}$. Once we establish the one-step bound in \eqref{eq:outOneStep}, to conclude the proof, we apply a lower bound on the number of updates performed by \texttt{DASA}, which is a function of $\tau_{avg}$ and that we obtain in Lemma~\ref{th:lemmaAtleast}. In the next paragraph, we provide the auxiliary lemmas needed to obtain the bounds illustrated above. 

\textbf{Auxiliary Lemmas.} In this paragraph, we will often use the following two basic results. For any $a, b \in \mathbb{R}, c\geq0$, and for $a_i\in\mathbb{R}, i = 1, ..., N$,
\begin{equation}\label{eq:basics}
\begin{aligned}
ab &= a\sqrt{c}\frac{b}{\sqrt{c}} \leq \frac{1}{2}\pars{ca^2 + \frac{b^2}{c}},\\
\pars{\sum_{i = 1}^{N}a_i}^2&\leq N\sum_{i = 1}^{N}a_i^2.
\end{aligned}
\end{equation}
For the purpose of the proof and without loss of generality, we define also samples and iterates for negative iterations, defining $\thet_j = \thet_0$ and $o_{i, j} = o_{i, 0}$ for $j\leq 0$ and for all $i = 1, ..., N$. We will also use the fact that from the definition of mixing time $\tmix$ (see Definition~\ref{def:mix}), we have
\begin{equation}\label{eq:mixing}
    \eta_{i,k}(\thet)\leq \alpha^2(\|\thet\|+\sigma).
\end{equation} 
We start the convergence analysis by proving a lemma, partially inspired by Lemma 4 in~\cite{cohen2021asynchronous}, that establishes a lower bound on the number of updates that the server performs when executing \texttt{DASA}, which is a function of the average delay $\tau_{avg}$.
\begin{lemma}\label{th:lemmaAtleast}
Let $\tau_{avg}$ be the average delay experienced by agents up to iterate $T$, $\tau_{avg}=\frac{1}{NT}\sum_{t=1}^{T}\sum_{i=1}^{N}\tau_{i,t}$. The number of updates that \texttt{DASA} makes is at least $\lceil{\frac{T}{8\tau_{avg}+8}}\rceil$.
\begin{proof}
    
Recall the definition of the average delay up to time $T$, $\tau_{avg}=\frac{1}{NT}\sum_{t=1}^{T}\sum_{i=1}^{N}\tau_{i,t}$.
Consider $U_{2\tau_{avg}}$, the number of steps $t$ for which the delay $\tau_t=\frac{1}{N}\sum_{i=1}^{N}\tau_{i,t}$ is at least $2\tau_{avg}$. We must have $U_{2\tau_{avg}} \leq \frac{T}{2}$ (otherwise the total sum of delays exceeds $\tau_{avg} T$, contradicting the definition of $\tau_{avg}$). On the other hand, let $m$ be the number of updates that the algorithm makes. Let $t_1 < t_2 < \ldots < t_m$ be the steps in which an update is made. Denote $t_0 = 0$ and $t_{m+1} = T$. Now, fix $i$ and consider the steps at times $s_n = t_i + n$ for $n \in [1, 2, \ldots, t_{i+1} - t_i - 1]$. In all those steps no update takes place and $\x_{s_n} = \x_{t_i}$. We must have $\tau_{i,s_n} > n$ for all $n$ for at least $\frac{N}{2}$ agents (otherwise $\x_k = \x_{t_{i,k}}$ for $k = s_n$ for more than $\frac{N}{2}$ agents and an update occurs since the median becomes zero). In particular, we have that $\tau_{i,s_n} \geq 4\tau_{avg}$ for at least $\frac{N}{2}$ of agents in at least $t_{i+1} - t_i - 1 - 4\tau_{avg}$ steps. Which means $\tau_{s_n}= \frac{1}{N}\sum_{i=1}^{N}\tau_{i,s_n}\geq 2\tau_{avg}$ in at least $t_{i+1} - t_i - 1 - 4\tau_{avg}$ steps. Formally,
\eqal{}{
    \text{\# of steps in $[t_i, t_{i+1}]$ with delay bigger or equal to $2\tau_{avg}$}\\\geq \max\{0,t_{i+1} - t_i - 1 - 4\tau_{avg}\}\\\geq t_{i+1} - t_i - 1 - 4\tau_{avg}.
}
Hence, $U_{2\tau_{avg}} \geq \sum_{i=0}^{m-1} (t_{i+1} - t_i - 1 - 4\tau_{avg})
= T - m(1 + 4\tau_{avg}).$
Finally, it follows that $T - m(1 + 4\tau_{avg}) \leq \frac{T}{2}$ which implies $m \geq \lceil\frac{T}{8(\tau_{avg}+1)}\rceil$.
\end{proof}
\end{lemma}
In the next lemma, we provide a bound on the expected squared norm of the \texttt{DASA} update direction, $\E{\|\vk\|^2}$, obtaining precisely what we outlined in~\eqref{eq:outDesired}.
\begin{lemma}\label{lemma:boundvk}
For $k\geq 0$, the following holds,
\begin{equation}\label{eq:lemmavk}
\begin{aligned}
\E{\|\vk\|^2}&\leq \bO{L^2}\E{\delta_k^2} + D, \hspace{0.2cm}\textrm{with}\\
D &\triangleq\bO{L^2\frac{\sigma^2}{N}} + \bO{ \sigma^2\alpha^{2}} +\bO{L^2\epsilon^2}.
\end{aligned}
\end{equation}
\begin{proof} We write
\eqal{eq:vkBound}{
\|\vk\|^2
&\leq \frac{12}{N^2}(V_1 + V_2+V_3), \hspace{0.3cm} \textrm{with}
\\V_1 & =  \|\sum_{i \in \calI}\bfg (\x_k,o_{i, t_{i,k}}) - \bfg(\xs, o_{i, t_{i,k}})\|^2,
\\V_2 & = \|\sum_{i \in \calI}\bfg(\xs, \odit)\|^2,
\\V_3 & = \|\sum_{i \in \calI}\dgit - \bfg(\x_k, o_{i, t_{i,k}})\|^2.
}
Note that the above bound holds regardless of $\calX_k = 0$ or $\calX_k = 1$.
We now bound $V_1$,
\eqal{}{
V_1 &\leq \frac{N}{2}\sum_{i \in \calI}\|\bfg (\x_k,o_{i, t_{i,k}}) - \bfg(\xs, o_{i, t_{i,k}})\|^2
\\&\overset{\eqref{eq:Lipschitz}}{\leq} \frac{N^2L^2 }{2}\delta_{k}^2. \hspace{0.3cm} \textrm{Thus,}
\\ \E{V_1} &\leq \frac{N^2L^2 }{2}\E{\delta_{k}^2}.
}
We now proceed to bound $V_2$,
\eqal{}{
V_2 &= V_{21} + V_{22}, \hspace{0.3cm} \textrm{with}
\\V_{21} &= \sum_{i\in \calI}\|\bfg(\xs, \odit)\|^2
\\V_{22} &= \sum_{\substack{i,j\in \calI\\i\neq j}}\langle \bfg(\xs, \odit),\bfg(\xs, \odjt) \rangle.
}
We see that, using \eqref{eq:bGSq}, we get
\eqal{}{
V_{21}&\leq 2L^2N(\|\xs\|^2 + \sigma^2)
\leq 4L^2N\sigma^2.
}
Now, using the fact that the observations $o_{i, k}$ and $o_{j, k'}$ are independent for $i\neq j$ and for any $k, k'\geq 0$,
\eqal{}{
\E{V_{22}} = \sum_{\substack{i,j\in \calI\\i\neq j}}&\langle \E{\E{\bfg(\xs, \odit)|o_{i, t_{i,k} - \tmix}}},
\\&\E{\E{\bfg(\xs, \odjt)|o_{j, t_{j,k} - \tmix}}} \rangle.
} 
Using the fact that $\barg(\xs) = 0$, and the Cauchy-Schwarz inequality followed by the Jensen's inequality, we can write
\eqal{}{
\E{V_{22}}&\leq \sum_{\substack{i,j\in \calI\\i\neq j}}\E{\mudiktau(\xs)}\times\E{\mudjktau(\xs)}
\\&
\leq \frac{1}{4}N^2\alpha^{4}(\|\xs\| + \sigma)^2\leq N^2\alpha^{4}\sigma^2,
}
where we used the mixing time definition and the consequent inequality~\eqref{eq:mixing}. For $V_3$ we have
\eqal{}{V_3 & \leq \frac{N}{2}\sum_{i \in \calI}\|\dgit - \bfg(\x_k, o_{i, t_{i,k}})\|^2,
\\&\leq \frac{N}{2}\sum_{i \in \calI}L^2\|\thet_{t_i,k} - \x_k\|^2,
\\&\leq \frac{N^2L^2\epsilon^2}{4}.}

Substituting the above bounds on $\E{V_1}$, $\E{V_2}$, and $\E{V_3}$ into \eqref{eq:vkBound}, we can conclude the proof of the lemma.
\end{proof}
\end{lemma}
We need the following result, which provides a bound to the expected drift term $\E{\delta_{k, \tmix}^2}$.
\begin{lemma}\label{lemma:drift}
    Let $D$ be as defined in~\eqref{eq:lemmavk}. There exists a constant $C'\geq 1$ such that for $\alpha\leq \frac{1}{C'L^2\tmix}$ we have
\begin{equation}\label{eq:driftAndBh}
\begin{aligned}
\E{\delta_{k, \tmix}^2}&\leq \bO{\alpha^2\tmix^2L^2}\E{\delta_k^2} + B_\tmix, \hspace{0.2cm}\textrm{with}\\
B_\tmix &= \bO{\alpha^2\tmix^2D}.
\end{aligned}
\end{equation}
\begin{proof}
We start out by establishing a bound on $\E{\delta_{k}^2}$ as a function of $\E{\delta_{k-\tmix}^2}$,
\begin{equation}
\begin{aligned}
{\delta_{k+1}^2} &= \calX_k\pars{{\delta_{k}^2} - 2\alpha{\inner{\vk}{\thet^* - \thet_k}} + \alpha^2\E{\|\vk\|^2}}\\
& + \pars{1-\calX_k}\delta_k^2,
\\&
\leq (1+\alpha)\delta_k^2 + 2\alpha\|\vk\|^2,
\end{aligned}
\end{equation}
where we used~\eqref{eq:basics}. Now taking the expectation and using Lemma~\ref{lemma:boundvk} we get
\begin{equation}
\begin{aligned}
\E{\delta_{k+1}^2}&\leq (1+\alpha)\E{\delta_k^2}+\bO{\alpha L^2}\E{\delta_k^2} +\alpha D\\
&\leq (1+\bO{\alpha L^2})\E{\delta_k^2} + \alpha D,
\end{aligned}
\end{equation}
where $D$ is defined as in~\eqref{eq:lemmavk}.
Iterating the above inequality, we get, for all $k'$ such that $k-\tmix\leq k'\leq k$,
\begin{equation}\label{eq:boundPastIterate}
\begin{aligned}
\E{\delta_{k'}^2}&\leq \pars{1+\bO{\alpha L^2}}^\tmix\E{\delta_{k-\tmix}^2}\\&+\alpha D\sum_{j = 0}^{\tmix-1}\pars{1+\bO{\alpha L^2}}^j
\\&
\leq 2\E{\delta_{k-\tmix}^2} + 2\alpha \tmix D,
\end{aligned}
\end{equation}
where we used the fact that, given a constant $C\geq 1$ large enough, $\pars{1+C\alpha L^2}^\tmix\leq \exp\pars{C\tmix\alpha L^2}\leq \exp\pars{0.25}\leq 2$ for $\alpha\leq\frac{1}{4C\tmix L^2}$.
Now, note that, for $k\geq2\tmix$, we can apply Lemma~\ref{lemma:boundvk}, getting
\begin{equation}
\begin{aligned}
\E{\delta_{k, \tmix}^2}&\leq \tmix\sum_{l= k-\tmix}^{k-1}\E{\|\thet_{l+1}-\thet_l\|^2}\\
&= \tmix\alpha^2\sum_{l=k-\tmix}^{k-1}\E{\|\mathbf{v}_l\|^2}
\\&
\overset{\eqref{eq:lemmavk}}{\leq} \tmix\alpha^2\sum_{l=k-\tmix}^{k-1}\bO{L^2}\E{\delta_l^2}+D
\\&
\leq  \bO{\alpha^2 \tmix^2D} \\&+ \bO{\tmix\alpha^2L^2}\sum_{l = k-\tmix}^{k-1}\E{\delta_l^2}.    
\end{aligned}
\end{equation}
Now, using~\eqref{eq:boundPastIterate} to upper bound $\E{\delta_l^2}$,
\begin{equation}
\begin{aligned}
\E{\delta_{k, \tmix}^2}& \leq  \bO{\alpha^2 \tmix^2D} +\bO{\tmix\alpha^2L^2}\\& \times\sum_{l = k-\tmix}^{k-1}\pars{\E{\delta_{k-\tmix}^2} +2\alpha \tmix D}
\\&
\leq \bO{\alpha^2\tmix^2L^2}\E{\delta_{k-\tmix}^2} \\&+ \bO{\alpha^2 \tmix^2D},
\end{aligned}
\end{equation}
where the last inequality holds by setting $\alpha\leq \frac{1}{C'L^2\tmix}$ for a constant $C'$ large enough.  Now, note that $\delta_{k-\tmix}^2\leq 2\delta_k^2 + 2\delta_{k, \tmix}^2$, from which we can conclude, setting $\alpha\leq\frac{1}{C'L^2\tmix}$, with $C'$ a constant large enough, noting that $B_{\tmix} = \bO{\alpha^2\tmix^2D}$.
\end{proof}
\end{lemma}
We will also need bounds on some further drift terms, that we group together in the following lemma.
\begin{lemma}
Let $\Tilde{\tau} \in \mathcal{T}_k$ and let $B_{\tmix}$ be as in~\eqref{eq:driftAndBh}. The following inequalities hold,
\begin{equation}\label{eq:boundsGrouped}
\begin{aligned}
(i)&\hspace{0.2cm} \E{\delta_{k-\Tilde{\tau}, \tmix}^2}&&\leq \bO{\alpha^2\tmix^2L^2}\E{\delta_{k}^2} \\&& &+ \bO{B_{\tmix}};\\[0.2cm]
(ii)&\hspace{0.2cm} \E{\delta_{k, \tmix+\Tilde{\tau}}^2} &&\leq \bO{\alpha^2\tmix^2L^2}\E{\delta_{k}^2} \\&& &+ \bO{B_{\tmix}} + \bO{\epsilon^2};\\[0.2cm]
(iii)&\hspace{0.2cm} \E{\delta_{k- \tmix-\Tilde{\tau}}^2} &&\leq \bO{\E{\delta_{k}^2}} \\&& &+ \bO{B_{\tmix}} + \bO{\epsilon^2}.
\end{aligned}
\end{equation}
\begin{proof}
$(i)$ Using Lemma~\ref{lemma:drift}, we can bound 
\begin{equation}
\begin{aligned}
\E{\delta_{k-\Tilde{\tau}, \tmix}^2} &\leq \bO{\alpha^2\tmix^2L^2}\E{\delta_{k-\Tilde{\tau}}^2} + B_{\tmix}
,\\&
\leq \bO{\alpha^2\tmix^2L^2}\E{\delta_k^2}
\\&+\bO{\alpha^2\tmix^2L^2\epsilon^2} + B_{\tmix}
,\\&
\leq \bO{\alpha^2\tmix^2L^2}\E{\delta_k^2} + \bO{B_{\tmix}}.
\end{aligned}
\end{equation}
For $(ii)$, note that, using the above inequality,
\begin{equation}
\begin{aligned}
\E{\delta_{k, \Tilde{\tau} + \tmix}^2} &\leq 2\E{\delta_{k, \Tilde{\tau}}^2} + 2\E{\delta_{k-\Tilde{\tau}, \tmix}^2},\\&\leq
2\epsilon^2 + \bO{\alpha^2\tmix^2L^2}\E{\delta_k^2}\\
&+ \bO{B_{\tmix}}.
\end{aligned}
\end{equation}
Finally, we can use the above inequality to prove $(iii)$,
\begin{equation}
\begin{aligned}
\E{\delta_{k-\tmix-\Tilde{\tau}}^2}&\leq 2\E{\delta_k^2} + 2\E{\delta_{k, \Tilde{\tau}+\tmix}^2}
,\\&\leq \bO{\E{\delta_k^2}} + \bO{\epsilon^2} + \bO{B_\tmix},
\end{aligned}
\end{equation}
where the last inequality follows for $\alpha\leq \frac{1}{\tmix L^2C}$, for some constant $C\geq1$.
\end{proof}
\end{lemma}
Equipped with the lemmas above, we can now focus on the more challenging part of the proof of Theorem~\ref{th:main}, which consists in bounding $\E{\psi_k}$ with a bound of the form we anticipated in the outline of the proof (see~\eqref{eq:outDesired}). We proceed in doing so by stating and proving the following lemma.
\begin{lemma}\label{lemma:psik}
Let $D$ be defined as in~\eqref{eq:lemmavk}. There exists a constant $C\geq 1$ such that, for $\alpha\leq \frac{1}{CL^2\tmix}$,
\begin{equation}
\begin{aligned}
\E{\psi_k}&\leq \bO{\alpha\tmix L^2}\E{\delta_k^2} + \bO{\alpha\tmix D}. 
\end{aligned}
\end{equation}
\begin{proof}
We start by defining $\barg_{N,k} \triangleq \frac{2}{N}\sum_{i\in\calI}\barg\pars{\thet_{t_{i,k}}}$, and, recalling the definition of $\psi_k$ in~\eqref{eq:outRecursion}, we write 
\begin{equation}
\begin{aligned}
\psi_k &= K_1 + K_2, \hspace{0.2cm}\textrm{with}\\
    K_1& = \inner{\barg(\thet_k) - \barg_{N, k}}{\thet_k-\xs}\\
    K_2& = \inner{\thet_k - \xs}{\barg_{N,k} - \vk}.
\end{aligned}
\end{equation}
We now bound $K_1$. Using~\eqref{eq:basics} and~\eqref{eq:Lipschitz},
\begin{equation}
\begin{aligned}
K_1&\leq \alpha\delta_k^2 + \bO{\frac{1}{\alpha N^2}}\left\Vert\sum_{i\in \calI}\barg(\thet_k) - \barg(\thet_{t_{i,k}})\right\Vert^2
\\&
\leq \alpha\delta_k^2 + \bO{\frac{L^2}{\alpha N}}\sum_{i\in \calI}\left\Vert\thet_k - \thet_{t_{i,k}}\right\Vert^2
\\&
\leq \alpha\delta_k^2 + \bO{L^2\frac{\epsilon^2}{\alpha}}
\\&
\leq \alpha\delta_k^2 + \bO{L^2\max\left\{\alpha^2, \frac{2\alpha}{N}\right\}}.
\end{aligned}
\end{equation}
We then proceed to bound $K_2$.
\begin{equation}
\begin{aligned}
K_2 &= K_{21} + K_{22} + K_{23}, \hspace{0.1cm}\textrm{with}
\\
K_{21} &= \inner{\thet_k - \xs}{\frac{2}{N}\sum_{i\in\calI}\barg\pars{\thet_{t_{i,k}}} - \barg\pars{\thet_{t_{i,k} - \tmix}}},\\
K_{22} &= \langle\thet_k - \xs, \\&\hspace{0.4cm}\frac{2}{N}\sum_{i\in\calI}\barg\pars{\thet_{t_{i,k} - \tmix}} - \bfg\pars{\thet_{t_{i,k}-\tmix}, o_{i,t_{i,k}}}\rangle,\\
K_{23} &= \langle\thet_k - \xs,\\&\hspace{0.4cm} \frac{2}{N}\sum_{i\in\calI}\bfg\pars{\thet_{t_{i,k} - \tmix}, o_{i,t_{i,k}}} - \bfg\pars{\thet_{t_{i,k}}, o_{i,t_{i,k}}}\rangle.
\end{aligned}
\end{equation}
Using~\eqref{eq:Lipschitz} and~\eqref{eq:basics}, $K_{21}$ and $K_{23}$ can be bounded in the same way, writing $K_{21} = \frac{2}{N}\sum_{i\in\calI}K_{21, i}$, with
\begin{equation}
\begin{aligned}
K_{21, i}& = \inner{\thet_k - \xs}{\barg\pars{\thet_{t_{i,k}}} - \barg\pars{\thet_{t_{i,k} - \tmix}}},
\end{aligned}
\end{equation}
and note that, recalling that $t_{i,k} = k-\tau_{i,k}$, and that $\tau_{i,k}\in\calT_k$, taking the expectation and using~\eqref{eq:boundsGrouped}-$(i)$, we obtain
\begin{equation}
\begin{aligned}
\E{K_{21,i}}&\leq \alpha\tmix\E{\delta_k^2} + \frac{L^2\E{\delta_{t_{i,k}, \tmix}^2}}{\alpha\tmix}
\\&
\leq \alpha\tmix\E{\delta_k^2} + \bO{\alpha\tmix L^2}\E{\delta_{k}^2}\\&+ \bO{\frac{B_{\tmix}}{\alpha\tmix}}
\\&
\leq \bO{\alpha\tmix L^2}\E{\delta_k^2} + \bO{\alpha\tmix D},
\end{aligned}
\end{equation}
and we can obtain the same bound for $\E{K_{23}}$. We now bound $\E{K_{22}}$. Let $\Hat{\tau}_k \triangleq \max\{\tau_{i,k}\in\calT_k\}$ and let
\begin{equation}
\Delta G_{i,k} \triangleq \barg\pars{\thet_{t_{i,k} - \tmix}} - \bfg\pars{\thet_{t_{i,k} -\tmix}, o_{i,t_{i,k}}}.
\end{equation}
Note that we can write $K_{22} = \bar{K}_1 + \bar{K}_2$, with
\begin{equation}
\begin{aligned}
\bar{K}_1& = \frac{2}{N}\sum_{i\in\calI}\langle \thet_{k-\tmix - \hat{\tau}_k}-\xs,\Delta G_{i,k} \rangle, \hspace{0.2cm}\\
\bar{K}_2& = \langle \thet_k-\thet_{k-\tmix - \hat{\tau}_k},\frac{2}{N}\sum_{i\in\calI}\Delta G_{i,k} \rangle.
\end{aligned}
\end{equation}
Now we write ${\bar{K}_1} = \frac{2}{N}\sum_{i\in\calI}\bar{K}_{1, i}$, with
\begin{equation}
\begin{aligned}
\bar{K}_{1, i}& = \langle \thet_{k-\tmix - \hat{\tau}_k}-\xs,\Delta G_{i,k} \rangle.
\end{aligned}
\end{equation}
Let ${t}_{mix}^{(k)} = k-\Hat{\tau}_k- \tmix$. Taking the expectation,
\begin{equation}
\begin{aligned}
\E{\bar{K}_{1, i}} = \E{\inner{\thet_{{t}_{mix}^{(k)}} - \xs}{\E{\Delta G_{i,k}|\thet_{{t}_{mix}^{(k)}}}}}.
\end{aligned}
\end{equation}
Now, define the two following sets,
\begin{equation}
\begin{aligned}
\Theta_{i,k} &= \{\thet_{t_{i,k} - \tmix}, o_{i,t_{i,k}-\tmix}\},
\\
\bar{\Theta}_{i,k}& = \{\thet_{t_{i,k} - \tmix}, o_{i,t_{i,k}-\tmix}, \thet_{{t}_{mix}^{(k)}}\}.
\end{aligned}
\end{equation}
Now note that from standard results in probability theory (see~\cite[Theorem~4.1.13.]{durrett2019probability}),
\begin{equation}\label{eq:firstCond}
\begin{aligned}
\E{\Delta G_{i,k}|\thet_{{t}_{mix}^{(k)}}} = \E{\E{\Delta G_{i,k}|\bar{\Theta}_k}|\thet_{{t}_{mix}^{(k)}}}.
\end{aligned}
\end{equation}
Now note further that the following equality holds:
\begin{equation}\label{eq:secondCond}
    \E{\Delta G_{i,k}|\Bar{\Theta}_{i,k}} = \E{\Delta G_{i,k}|{\Theta}_{i,k}}.
\end{equation}
Indeed, since for $i\in\calI_k$, $\hat{\tau}_k\geq \tau_{i,k}$, we have ${t}_{mix}^{(k)}\leq t_{i,k} - \tmix$. Also note that the parameter $\thet_{{t}_{mix}^{(k)}}$ is a function of the samples of the Markov chain $o_{i, l}$ only for $l\leq {t}_{mix}^{(k)}-1<t_{i,k} - \tmix$. Hence, thanks to the Markov property, once we condition on $o_{i,t_{i,k} - \tmix}$, the random variable $\Delta G_{i,k}$ does not depend on the statistical information contained in $\thet_{{t}_{mix}^{(k)}}$. In addition, from the definition of mixing time $\tmix$ (see Definition~\ref{def:mix}) the following inequality holds:
\begin{equation}\label{eq:thirdCond}
\begin{aligned}
\left\Vert \E{\Delta G_{i,k}|{\Theta}_{i,k}}\right\Vert\leq\alpha^2(\|\thet_{t_{i,k} - \tmix}\| + \sigma). 
\end{aligned}
\end{equation}
Using these last facts contained in equations~\eqref{eq:firstCond},~\eqref{eq:secondCond} and~\eqref{eq:thirdCond}, we can now proceed to bound $\E{\bar{K}_{1, i}}$ as follows:
\begin{equation}
\begin{aligned}
\E{\bar{K}_{1, i}}&= \E{\inner{\thet_{{t}_{mix}^{(k)}} - \xs}{\E{\Delta G_{i,k}|\thet_{{t}_{mix}^{(k)}}}}}
\\&
\overset{\eqref{eq:firstCond}, \eqref{eq:secondCond}}{=}\E{\inner{\thet_{{t}_{mix}^{(k)}} - \xs}{\E{\E{\Delta G_{i,k}|{\Theta}_k}|\thet_{{t}_{mix}^{(k)}}}}} 
\\&
\leq\E{\delta_{t_{mix}^{(k)}}\left\Vert \E{\E{\Delta G_{i,k}|{\Theta}_k}|\thet_{{t}_{mix}^{(k)}}} \right\Vert}
\\&
\overset{(a)}{\leq} \E{\delta_{t_{mix}^{(k)}}\E{\left\Vert \E{\Delta G_{i,k}|{\Theta}_k} \right\Vert}|\thet_{{t}_{mix}^{(k)}}}
\\&
\overset{\eqref{eq:thirdCond}}{\leq}\E{\delta_{t_{mix}^{(k)}}\E{\alpha^2\pars{\|\thet_{t_{i,k}-\tmix}\| + \sigma}|\thet_{{t}_{mix}^{(k)}}}}
\\&
\overset{(b)}{\leq}\E{\delta_{t_{mix}^{(k)}}\alpha^2\pars{\|\thet_{t_{i,k}-\tmix}\| + \sigma}}
\end{aligned}
\end{equation}
where for $(a)$ we used Jensen's inequality for conditional expectation (see~\cite[Theorem 4.1.10.]{durrett2019probability}) and $(b)$ follows from the fact that, for random variables $X$ and $Y$, $\E{XY|X} = X\E{Y|X}$~\cite[see Theorem 4.1.14.]{durrett2019probability}.
We can now bound $\E{\bar{K}_1}$ using~\eqref{eq:basics} and~\eqref{eq:boundsGrouped}, recalling that $B_{\tmix} = \bO{\alpha^2\tmix^2D}$, as follows:
\begin{equation}
\begin{aligned}
\E{\bar{K}_{1, i}}&\leq \E{\delta_{t_{mix}^{(k)}}\alpha^2\pars{\|\thet_{t_{i,k}-\tmix}\| + \sigma}}
\\&\overset{\eqref{eq:basics}}{\leq} \E{\alpha\delta_{t_{mix}^{(k)}}^2 + \alpha^{3}\bO{\delta_{t_{i,k} - \tmix}^2 + \sigma^2}}
\\&
\overset{\eqref{eq:boundsGrouped}}{\leq} \bO{\alpha\E{\delta_k^2}} + \bO{\alpha D},
\end{aligned}
\end{equation}
where the last inequality follows because $\alpha\leq \frac{1}{CL^2\tmix}$. We can then conclude that $\E{\bar{K}_1}\leq \bO{\alpha\E{\delta_k^2}} + \bO{\alpha D}$. Now we proceed to bound $\bar{K}_2$. Using~\eqref{eq:basics},
\begin{equation}
\begin{aligned}
\E{\bar{K}_2}&\leq \frac{1}{2\alpha\tmix}\E{\delta_{k, \tmix+\Tilde{\tau}}^2} \\&+ {\bO{\frac{\alpha\tmix}{N^2}}}\E{\left\Vert\sum_{i\in\calI}\Delta G_{i,k}\right\Vert^2}.
\end{aligned}
\end{equation}
Using~\eqref{eq:boundsGrouped}-$(ii)$, we can get, recalling the choice of $\epsilon$ in~\eqref{eq:parameters},
\begin{equation}
\begin{aligned}
\frac{1}{2\alpha\tmix}\E{\delta_{k, \tmix+\Tilde{\tau}}^2}&\leq \bO{\alpha\tmix L^2}\E{\delta_{k}^2}
\\&
+\bO{\frac{B_{\tmix}}{\alpha\tmix}} + \bO{\frac{\epsilon^2}{\alpha\tmix}}
\\&
\leq \bO{\alpha\tmix L^2}\E{\delta_{k}^2}
\\&
+\bO{\alpha\tmix D}.
\end{aligned}
\end{equation}
Finally, we proceed to bound $\E{\left\Vert\sum_{i\in\calI}\Delta G_{i,k}\right\Vert^2}$:
\begin{equation}
\begin{aligned}
\E{\left\Vert\sum_{i\in\calI}\Delta G_{i,k}\right\Vert^2}&\overset{\eqref{eq:basics}}{\leq} \E{G_1} + \E{G_2}\hspace{0.2cm}\textrm{with} 
\\G_1&=2\left\Vert\sum_{i\in\calI}\barg(\thet_{t_{i,k} - \tmix})\right\Vert^2
\\
G_2&= 2\left\Vert\sum_{i\in\calI}\bfg\pars{\thet_{t_{i,k} -\tmix}, o_{i,t_{i,k}}}\right\Vert^2.
\end{aligned}
\end{equation}
Note that, using~\eqref{eq:strongConvex} and\eqref{eq:boundsGrouped}-$(iii)$, we have
\begin{equation}
\begin{aligned}
\E{G_1}&\overset{\eqref{eq:strongConvex}}{\leq} \bO{N^2\E{\delta_{t_{i,k}-\tmix}^2}}
\\&\overset{\eqref{eq:boundsGrouped}}{\leq} \bO{N^2\pars{{\E{\delta_{k}^2}} + {B_{\tmix}} + {\epsilon^2}}}.
\end{aligned}
\end{equation}
Note that $G_2$ can be bounded in the same way in which we bounded $V_2$ in Lemma~\ref{lemma:boundvk}, obtaining 
\begin{equation}
\E{G_2}\leq \bO{N^2}\pars{L^2\E{\delta_k^2} + \alpha^4\sigma^2 +L^2\frac{\sigma^2}{N}}.
\end{equation}
Substituting the bounds obtained on $G_1$ and $G_2$ we can conclude the bound on $\E{\bar{K}_2}$, obtaining
\begin{equation}
\E{\bar{K}_2}\leq \bO{\alpha\tmix L^2}\E{\delta_{k}^2}
+\bO{\alpha\tmix D}.
\end{equation}
With this last result, we conclude the proof of the lemma.
\end{proof}
\end{lemma}
\textbf{Conclusion of the proof.}
Equipped with the results of the previous paragraph, we can now conclude the proof of Theorem~\ref{th:main}. Note that Lemma~\ref{lemma:boundvk} and Lemma~\ref{lemma:psik} provide the desired bounds on $\E{\|\vk\|^2}$ and $\E{\psi_k}$ that we had showcased in the outline of the proof, in~\eqref{eq:outDesired}. If we substitute these bounds into~\eqref{eq:outRecursion}, we obtain the desired one-step improvement of~\eqref{eq:outOneStep}, by choosing $\alpha\leq\frac{\mu}{CL^2\tmix}$, for $C\geq1$ large enough. Now let $T\geq0$ be such that $I_T = 1$. Note that, thanks to Lemma~\ref{th:lemmaAtleast}, we know that the number of updates $l$ performed by \texttt{DASA} is $l\geq\frac{T}{8\pars{\tau_{avg} + 1}}$. Now noting that for $0\leq k\leq T-1$ the error $\E{\delta_{k+1}^2}$ either satisfies the one-step improvement in~\eqref{eq:outOneStep} or it is equal to $\E{\delta_{k}^2}$, we see that, unrolling the one-step improvement inequality in~\eqref{eq:outOneStep},
\begin{equation}
\begin{aligned}
\E{\delta_{T}^2}&\leq \exp\left({-\frac{\alpha\mu T}{8\pars{\tau_{avg} + 1}}}\right)\sigma^2 + \gamma,\hspace{0.2cm}\textrm{with}
\\[0.2cm]
\gamma& = \bO{\frac{\alpha\tmix\sigma^2L^2}{\mu}}\pars{\frac{1}{N} + \alpha}.
\end{aligned}
\end{equation}
For a more technical description of the derivation of the above inequality, the interested reader can check the analogous derivation of Eq.~(253) in Appendix~B.3.5 of~\cite{adibi2024stochastic}.
\section{Experiments}\label{sec:sim}
\begin{figure}[t]
\center
\includegraphics[width=1\columnwidth, trim ={2.7cm 0cm 3.3cm 0cm}, clip]{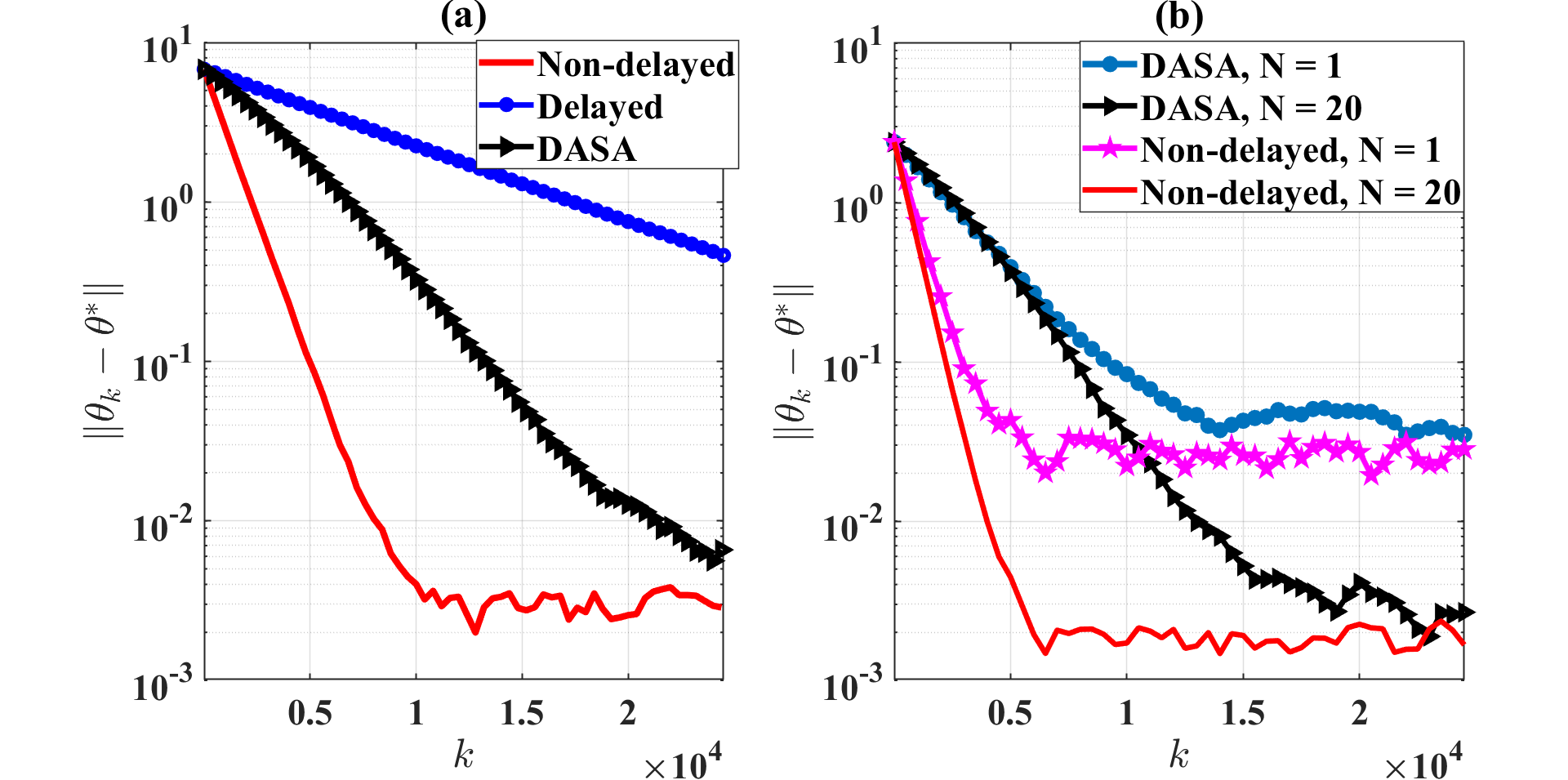}
	\caption{In (a), we show the superior performance of \texttt{DASA} compared to non-adaptive distributed SA under delayed updates (``Delayed") when the number of agents is $N = 10$. In (b), we show the convergence speedup effect comparing a single-agent ($N = 1$) and a multi-agent ($N = 20$) setting. In all simulations, we set $\tau_{max} = 50$.}
\label{fig:sim_results}
\end{figure}

In this section, we provide simulation results to validate our theoretical analysis and show the superior performance of \texttt{DASA} when compared to a non-adaptive baseline of asynchronous SA. We consider a simple instance of linear SA, i.e., temporal difference (TD) learning with linear function approximation. The setting that we consider is similar to the one adopted for the simulations in~\cite{han, dal2023federated} of federated TD learning, with the difference that we let the agents' TD update directions be subject to delays, i.e., the server updates the TD approximation parameter with $\vk$ defined in~\eqref{eq:algo}, where $\bfg\pars{\thet_{t_{i,k}}, o_{t_{i,k}}}$ is the TD direction computed by agent $i$ at (delayed) time $t_{i,k}$. For the TD learning setting, the SA approximation setting consists of estimating the best approximation $\thet^*$  of a value function associated to a policy $\mu$ that induces a Markov chain when applied to a Markov Decision Process (MDP) $\calM  = (\mathcal{S}, \mathcal{A}, \mathcal{P}, \mathcal{R}, \gamma)$, where we consider a finite state space $\mathcal{S}$ of dimension $n = 30$. With respect to the other notation, $\mathcal{A}$ is a finite set of actions, $\mathcal{P}$ is the set of action-dependent Markov transition kernels, $\mathcal{R}$ is a reward function, and the discount factor is $\gamma = 0.5$. We approximate the value function with a set of $r = 10$ basis vectors. For the considered TD learning setting, we have $\mu = (1-\gamma)\omega$, where $\omega>0$ is an eigenvalue that we compute explicitly to set the step sizes. We set the maximum delay to $\tau_{max} = 50$. In Figure~\ref{fig:sim_results}-(a), we consider a distributed SA setting with $N = 10$ agents and compare \texttt{DASA} against (i) a version of distributed SA with no delays, which we refer to as ``Non-delayed", and (ii) a non-adaptive version of distributed asynchronous SA, which we simply call ``Delayed". Convergence guarantees for ``Delayed" (in a single-agent setting) have been studied in~\cite{adibi2023min}, where it was shown that tight convergence bounds are achievable for a step size $\alpha$ inversely proportional to $\tmix\pars{\alpha} + \tau_{max}$. 
The plot clearly shows the advantage of the delay-adaptive solution (\texttt{DASA}) when compared to the non-adaptive SA algorithm: for the same number of updates, the approximation error achieved by \texttt{DASA} is almost two orders of magnitude smaller than that of the non-adaptive algorithm (``Delayed"). We can also notice how the approximation error to which \texttt{DASA} converges is quite close to the approximation error of the ``Non-delayed" version. In Figure~\ref{fig:sim_results}-(b), we illustrate the linear speedup effect that we established theoretically for \texttt{DASA} comparing a single-agent configuration ($N = 1$) to a multi-agent setting ($N = 20$). We include also the performance of the ``Non-delayed" version of the algorithm for the same numbers of agents $N = 1, 20$. With the same step size, when $N = 20$ agents are estimating the value function in parallel, the approximation error is improved by more than one order of magnitude, which validates the convergence speedup effect. We can indeed note that the speedup obtained by \texttt{DASA} has a similar performance to the speedup obtained for distributed SA without delays.
\section{Conclusion and Future Work}
We introduced \texttt{DASA}, a novel algorithm that effectively mitigates the effect of asynchronous and potentially unbounded network delays and ensures collaborative speedups in multi-agent SA. Future works include the design of more advanced delay-adaptive algorithms, extensions to fully distributed (peer-to-peer) network architectures and further empirical evaluations.
\bibliographystyle{IEEEtran} 
\bibliography{refs}

\begin{thebibliography}{10}
\providecommand{\url}[1]{#1}
\csname url@samestyle\endcsname
\providecommand{\newblock}{\relax}
\providecommand{\bibinfo}[2]{#2}
\providecommand{\BIBentrySTDinterwordspacing}{\spaceskip=0pt\relax}
\providecommand{\BIBentryALTinterwordstretchfactor}{4}
\providecommand{\BIBentryALTinterwordspacing}{\spaceskip=\fontdimen2\font plus
\BIBentryALTinterwordstretchfactor\fontdimen3\font minus
  \fontdimen4\font\relax}
\providecommand{\BIBforeignlanguage}[2]{{%
\expandafter\ifx\csname l@#1\endcsname\relax
\typeout{** WARNING: IEEEtran.bst: No hyphenation pattern has been}%
\typeout{** loaded for the language `#1'. Using the pattern for}%
\typeout{** the default language instead.}%
\else
\language=\csname l@#1\endcsname
\fi
#2}}
\providecommand{\BIBdecl}{\relax}
\BIBdecl

\bibitem{khodadadian}
S.~Khodadadian, P.~Sharma, G.~Joshi, and S.~T. Maguluri, ``Federated
  reinforcement learning: Linear speedup under {Markovian} sampling,'' in
  \emph{Proceedings of International Conference on Machine Learning}.\hskip 1em
  plus 0.5em minus 0.4em\relax PMLR, 2022, pp. 10\,997--11\,057.

\bibitem{dal2023federated}
N.~Dal~Fabbro, A.~Mitra, and G.~J. Pappas, ``Federated {TD}-learning over
  finite-rate erasure channels: Linear speedup under {Markovian} sampling,''
  \emph{IEEE Control Systems Letters}, 2023.

\bibitem{zhang2024finite}
C.~Zhang, H.~Wang, A.~Mitra, and J.~Anderson, ``Finite-time analysis of
  on-policy heterogeneous federated reinforcement learning,'' \emph{arXiv
  preprint arXiv:2401.15273}, 2024.

\bibitem{bertsekas1989convergence}
D.~P. Bertsekas and J.~N. Tsitsiklis, ``Convergence rate and termination of
  asynchronous iterative algorithms,'' in \emph{Proceedings of the 3rd
  International Conference on Supercomputing}, 1989, pp. 461--470.

\bibitem{koloskova2022sharper}
A.~Koloskova, S.~U. Stich, and M.~Jaggi, ``Sharper convergence guarantees for
  asynchronous {SGD} for distributed and federated learning,'' \emph{Advances
  in Neural Information Processing Systems}, vol.~35, pp. 17\,202--17\,215,
  2022.

\bibitem{stich2020error}
S.~U. Stich and S.~P. Karimireddy, ``The error-feedback framework: Better rates
  for {SGD} with delayed gradients and compressed updates,'' \emph{Journal of
  Machine Learning Research}, vol.~21, no.~1, pp. 9613--9648, 2020.

\bibitem{adibi2023min}
A.~Adibi, A.~Mitra, and H.~Hassani, ``Min-max optimization under delays,''
  \emph{arXiv preprint arXiv:2307.06886}, 2023.

\bibitem{AMARL_th}
H.~Shen, K.~Zhang, M.~Hong, and T.~Chen, ``Towards understanding asynchronous
  advantage actor-critic: Convergence and linear speedup,'' \emph{IEEE
  Transactions on Signal Processing}, vol.~71, pp. 2579--2594, 2023.

\bibitem{feyzmahdavian2014delayed}
H.~R. Feyzmahdavian, A.~Aytekin, and M.~Johansson, ``A delayed proximal
  gradient method with linear convergence rate,'' in \emph{2014 IEEE
  International Workshop on Machine Learning for Signal Processing
  (MLSP)}.\hskip 1em plus 0.5em minus 0.4em\relax IEEE, 2014, pp. 1--6.

\bibitem{adibi2024stochastic}
A.~Adibi, N.~Dal~Fabbro, L.~Schenato, S.~Kulkarni, H.~V. Poor, G.~J. Pappas,
  H.~Hassani, and A.~Mitra, ``Stochastic approximation with delayed updates:
  Finite-time rates under markovian sampling,'' in \emph{International
  Conference on Artificial Intelligence and Statistics}.\hskip 1em plus 0.5em
  minus 0.4em\relax PMLR, 2024, pp. 2746--2754.

\bibitem{bhandari2018}
J.~Bhandari, D.~Russo, and R.~Singal, ``A finite time analysis of temporal
  difference learning with linear function approximation,'' in
  \emph{Proceedings of Conference on learning theory}, 2018.

\bibitem{srikant2019finite}
R.~Srikant and L.~Ying, ``Finite-time error bounds for linear stochastic
  approximation and {TD} learning,'' in \emph{Proceedings of Conference on
  Learning Theory}.\hskip 1em plus 0.5em minus 0.4em\relax PMLR, 2019, pp.
  2803--2830.

\bibitem{chen2022finite}
Z.~Chen, S.~Zhang, T.~T. Doan, J.-P. Clarke, and S.~T. Maguluri,
  ``Finite-sample analysis of nonlinear stochastic approximation with
  applications in reinforcement learning,'' \emph{Automatica}, vol. 146, p.
  110623, 2022.

\bibitem{borkar2009stochastic}
V.~S. Borkar, \emph{{Stochastic Approximation: A Dynamical Systems
  Viewpoint}}.\hskip 1em plus 0.5em minus 0.4em\relax Springer, 2009, vol.~48.

\bibitem{meyn2023stability}
S.~Meyn, ``Stability of {Q-Learning} through design and optimism,'' \emph{arXiv
  preprint arXiv:2307.02632}, 2023.

\bibitem{cohen2021asynchronous}
A.~Cohen, A.~Daniely, Y.~Drori, T.~Koren, and M.~Schain, ``Asynchronous
  stochastic optimization robust to arbitrary delays,'' \emph{Advances in
  Neural Information Processing Systems}, vol.~34, pp. 9024--9035, 2021.

\bibitem{durrett2019probability}
R.~Durrett, \emph{{Probability: Theory and Examples}}.\hskip 1em plus 0.5em
  minus 0.4em\relax Cambridge university press, 2019, vol.~49.

\bibitem{han}
H.~Wang, A.~Mitra, H.~Hassani, G.~J. Pappas, and J.~Anderson, ``Federated
  temporal difference learning with linear function approximation under
  environmental heterogeneity,'' \emph{Transactions on Machine Learning
  Research}, 2023.

\end{thebibliography}
\end{document}